\documentclass{article}

\usepackage{hyperref}

\usepackage[accepted]{icml2023}  %

\usepackage{times}
\usepackage{latexsym}
\usepackage{amsmath}

\usepackage[T1]{fontenc}
\usepackage[utf8]{inputenc}

\usepackage{microtype}
\usepackage{bbding}
\usepackage{color,xcolor}
\usepackage{epsfig}
\usepackage{graphicx}

\usepackage{adjustbox}
\usepackage{array}
\usepackage{booktabs}
\usepackage{colortbl}
\usepackage{wrapfig}
\usepackage{hhline}
\usepackage{multirow}
\usepackage{subcaption} %

\usepackage{amsmath,amsfonts,amssymb}
\usepackage{bm}
\usepackage{nicefrac}
\usepackage{microtype}
\usepackage{mathtools}

\usepackage{changepage}
\usepackage{extramarks}
\usepackage{fancyhdr}
\usepackage{lastpage}
\usepackage{setspace}
\usepackage{soul}
\usepackage{xspace}

\usepackage{url}

\usepackage{enumerate}
\usepackage{enumitem}  %
\usepackage{titlesec}

\usepackage{makecell}

\usepackage{pifont} %

\usepackage{amsthm}
\usepackage{float}

\newcommand{\sect}[1]{Sec.~\ref{#1}}
\newcommand{\sectapp}[1]{Appendix~\ref{#1}}

\newcommand{\eqn}[1]{Equation~\ref{#1}}
\newcommand{\fig}[1]{Figure~\ref{#1}}

\newcommand{\tbl}[1]{Table~\ref{#1}}

\newcommand{\ignore}[1]{}

\renewcommand*{\thefootnote}{\fnsymbol{footnote}}

\DeclareMathAlphabet{\mathbfit}{OML}{cmm}{b}{it}

\makeatletter
\DeclareRobustCommand\onedot{\futurelet\@let@token\@onedot}
\def\@onedot{\ifx\@let@token.\else.\null\fi\xspace}

\def\eg{e.g\onedot} 
\def\ie{i.e\onedot} 
 
\def\etc{etc\onedot}

\def\method{SmoothQuant\xspace}

\newcommand{\myparagraph}[1]{\vspace{-6pt}\paragraph{#1}}

\usepackage{amsmath,amsfonts,bm}

\def\eqref#1{equation~\ref{#1}}

\def\1{\bm{1}}

\DeclareMathAlphabet{\mathsfit}{\encodingdefault}{\sfdefault}{m}{sl}
\SetMathAlphabet{\mathsfit}{bold}{\encodingdefault}{\sfdefault}{bx}{n}

\icmltitlerunning{SmoothQuant: Accurate and Efficient Post-Training Quantization for Large Language Models}

\begin{document}

\twocolumn[
\icmltitle{SmoothQuant: Accurate and Efficient \\ Post-Training Quantization for Large Language Models}

\icmlsetsymbol{equal}{*}

\begin{icmlauthorlist}
\icmlauthor{Guangxuan Xiao}{equal,mit}
\icmlauthor{Ji Lin}{equal,mit}
\icmlauthor{Mickael Seznec}{nv}
\icmlauthor{Hao Wu}{nv}
\icmlauthor{Julien Demouth}{nv}
\icmlauthor{Song Han}{mit}
\\
\url{https://github.com/mit-han-lab/smoothquant}
\end{icmlauthorlist}

\icmlaffiliation{mit}{Massachusetts Institute of Technology}
\icmlaffiliation{nv}{NVIDIA}

\icmlcorrespondingauthor{Guangxuan Xiao}{xgx@mit.edu}
\icmlcorrespondingauthor{Ji Lin}{jilin@mit.edu}

\icmlkeywords{Machine Learning, ICML}

\vskip 0.3in
]

\newcommand{\printAffiliations}[1]{%
\stepcounter{@affiliationcounter}%
{\let\thefootnote\relax\footnotetext{\hspace*{-\footnotesep}\ifdefined\isaccepted #1\fi%
\forloop{@affilnum}{1}{\value{@affilnum} < \value{@affiliationcounter}}{
\textsuperscript{\arabic{@affilnum}}\ifcsname @affilname\the@affilnum\endcsname%
\csname @affilname\the@affilnum\endcsname%
\else
{\bf AUTHORERR: Missing \textbackslash{}icmlaffiliation.}
\fi
}.
\ifdefined\icmlcorrespondingauthor@text
Correspondence to: \icmlcorrespondingauthor@text.
\else
{\bf AUTHORERR: Missing \textbackslash{}icmlcorrespondingauthor.}
\fi
}
}
}
\printAffiliationsAndNotice{\icmlEqualContribution} %

\begin{abstract}
    Large language models (LLMs) show excellent performance but are compute- and memory-intensive.
    Quantization can reduce memory and accelerate inference.
    However, existing methods cannot maintain accuracy and hardware efficiency at the same time.
    We propose \method, a training-free, accuracy-preserving, and general-purpose post-training quantization (PTQ) solution to enable 8-bit weight, 8-bit activation (W8A8) quantization for LLMs.
    Based on the fact that weights are easy to quantize while activations are not, \method smooths the activation outliers by offline \textit{migrating} the quantization difficulty from activations to weights with a mathematically equivalent transformation.
    \method enables an INT8 quantization of \textit{both} weights and activations for all the matrix multiplications in LLMs, including OPT, BLOOM, GLM, MT-NLG, Llama-1/2, Falcon, Mistral, and Mixtral models. 
    We demonstrate up to 1.56$\times$ speedup and 2$\times$ memory reduction for LLMs with negligible loss in accuracy.
    \method enables serving 530B LLM within a single node. Our work offers a turn-key solution that reduces hardware costs and democratizes LLMs. 
\end{abstract}

\section{Introduction}
Large-scale language models (LLMs) show excellent performance on various tasks~\citep{gpt3,opt}. However, serving LLMs is budget and energy-consuming due to their gigantic model size. For example, the GPT-3~\citep{gpt3} model contains 175B parameters, which will consume at least 350GB of memory to store and run in FP16, requiring 8$\times$48GB A6000 GPUs or 5$\times$80GB A100 GPUs just for inference.
Due to the huge computation and communication overhead, the inference latency may also be unacceptable to real-world applications.
\emph{Quantization} is a promising way to reduce the cost of LLMs~\cite{dettmers2022llmint8, zeroquant}. By quantizing the \emph{weights and activations} with low-bit integers, we can reduce GPU memory requirements, in size and bandwidth, and accelerate compute-intensive operations (\ie, \texttt{GEMM}\footnote{General matrix multiply} in linear layers, \texttt{BMM}\footnote{Batch matrix multiply} in attention). For instance, INT8 quantization of weights and activations can halve the GPU memory usage and nearly double the throughput of matrix multiplications compared to FP16.

\begin{figure}
    \centering
    \includegraphics[width=0.9\linewidth]{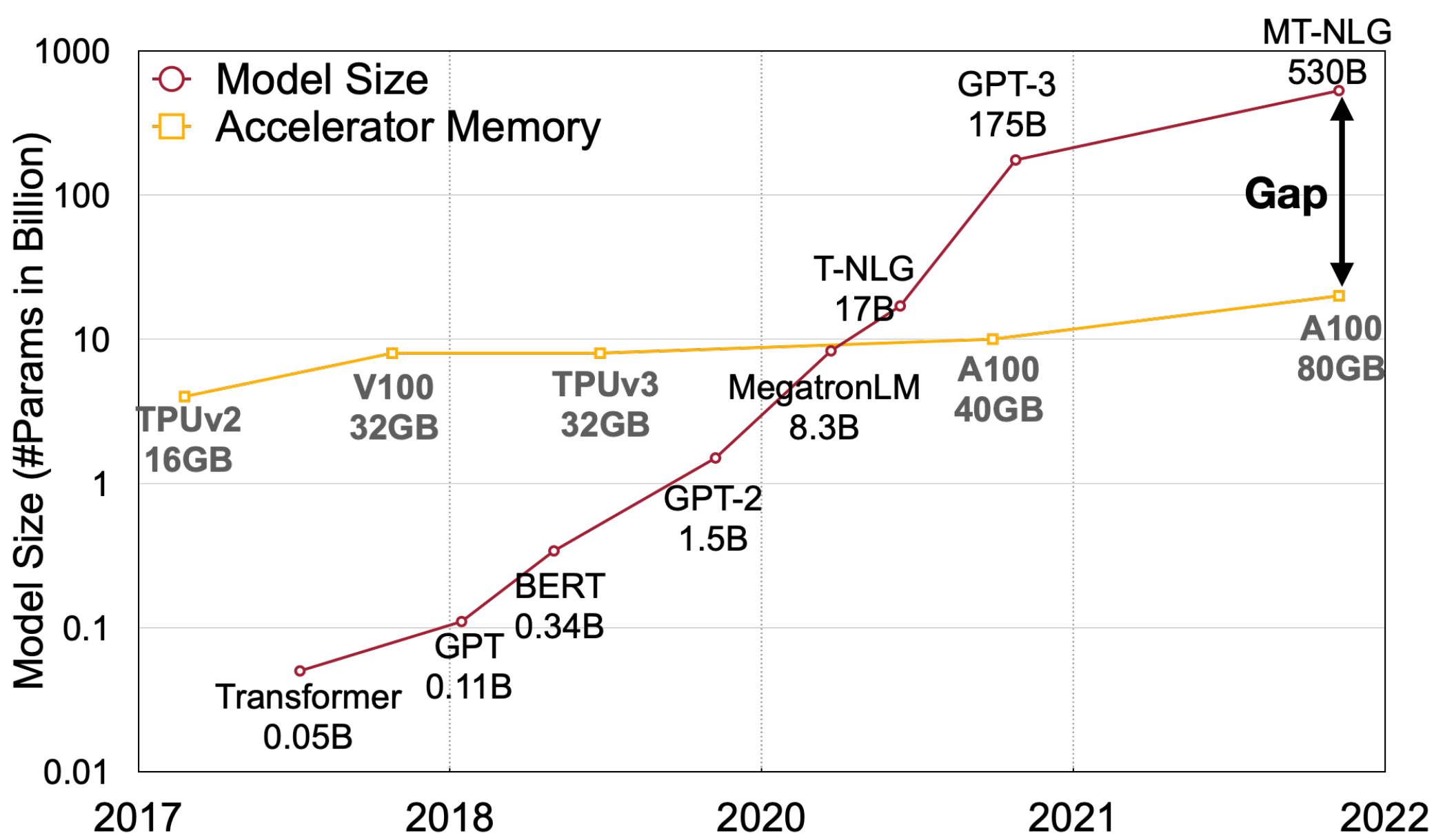}
    \caption{The model size of large language models is developing at a faster pace than the GPU memory in recent years, leading to a big gap between the supply and demand for memory. Quantization and model compression techniques can help bridge the gap. }
    \vspace{-1em}
    \label{fig:trend}
\end{figure}

However, unlike CNN models or smaller transformer models like BERT~\cite{bert}, the \emph{activations} of LLMs are difficult to quantize. When we scale up LLMs beyond 6.7B parameters, systematic outliers with large magnitude will emerge in activations~\cite{dettmers2022llmint8}, leading to large quantization errors and accuracy degradation.
ZeroQuant~\citep{zeroquant} applies dynamic per-token activation quantization and group-wise weight quantization (defined in \fig{fig:quantize_notation} \sect{sec:quantization}). It can be implemented efficiently and delivers good accuracy for GPT-3-350M and GPT-J-6B. However,  it can not maintain the accuracy for the large OPT model with 175 billion parameters (see Section~\ref{sec:accurate-quantization}).
\texttt{LLM.int8()}~\cite{dettmers2022llmint8} addresses that accuracy issue by further introducing a mixed-precision decomposition (\ie, it keeps outliers in FP16 and uses INT8 for the other activations). However, it is hard to implement the decomposition efficiently on hardware accelerators.
Therefore, deriving an \emph{efficient}, \emph{hardware-friendly}, and preferably \emph{training-free} quantization scheme for LLMs that would use INT8 for all the compute-intensive operations remains an open challenge.

We propose \method, an accurate and efficient post-training quantization (PTQ) solution for LLMs.
\method relies on a key observation: even if activations are much harder to quantize than weights due to the presence of outliers~\cite{dettmers2022llmint8},  different tokens exhibit similar variations across their channels.
\begin{figure}
    \centering
    \includegraphics[width=0.88\linewidth]{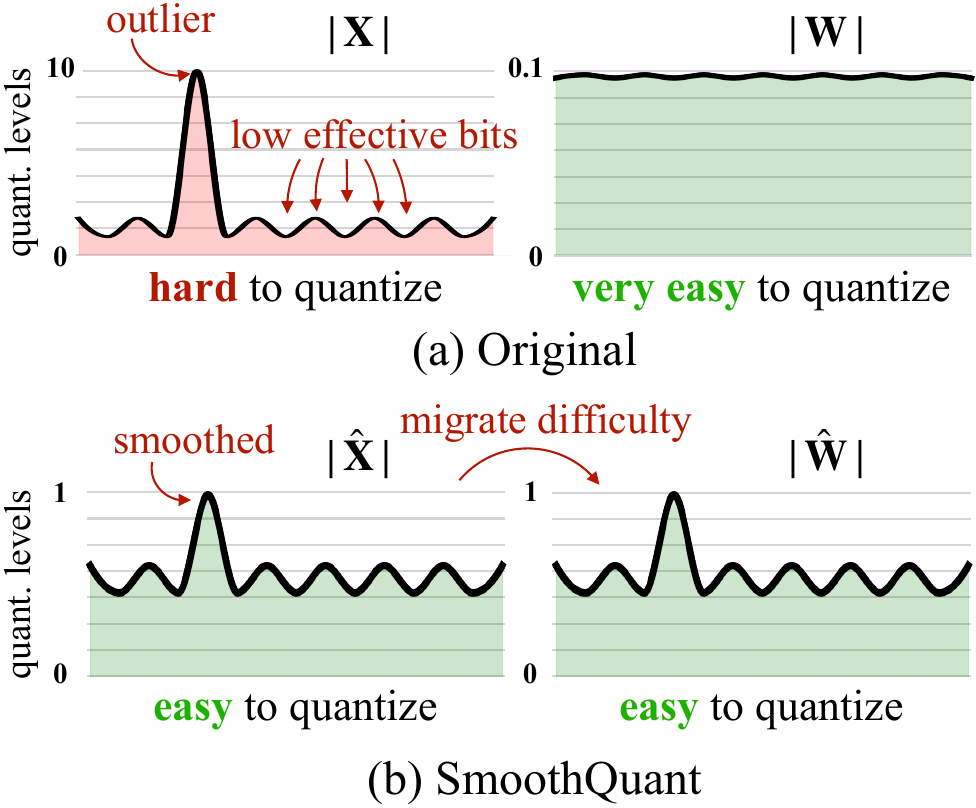}
    \caption{\method's intuition: the activation $\mathbf{X}$ is hard to quantize because outliers stretch the quantization range, leaving few effective bits for most values. We migrate the scale variance from activations to weights $\mathbf{W}$ during offline to reduce the quantization difficulty of activations. The smoothed activation $\hat{\mathbf{X}}$ and the adjusted weight  $\hat{\mathbf{W}}$ are both easy to quantize. 
       }
    \label{fig:intuition}
    \vspace{-10pt}
\end{figure}

Based on this observation, \method offline migrates the quantization difficulty from activations to weights (\fig{fig:intuition}).
\method proposes a mathematically equivalent per-channel scaling transformation that significantly smooths the magnitude across the channels, making the model quantization-friendly.
Since \method is compatible with various quantization schemes, we implement three efficiency levels of 
quantization settings for \method (see \tbl{tab:settings}, O1-O3).
Experiments show that \method is hardware-efficient: it can maintain the performance of OPT-175B~\cite{opt}, BLOOM-176B~\cite{scao2022bloom} , GLM-130B~\cite{zeng2022glm}, and MT-NLG 530B~\cite{smith2022using},
leading to up to 1.51$\times$ speed up and 1.96$\times$ memory saving on PyTorch.
\method is easy to implement. We integrate \method into FasterTransformer, the state-of-the-art transformer serving framework, achieving up to 1.56$\times$ speedup and halving the memory usage compared with FP16.
Remarkably, \method allows serving large models like OPT-175B using only half number of GPUs compared to FP16 while being faster, and enabling the serving of a 530B model within one 8-GPU node.
Our work democratizes the use of LLMs by offering a turnkey solution to reduce the serving cost. We hope \method can inspire greater use of LLMs in the future.

\section{Preliminaries}
\label{sec:quantization}
\textbf{Quantization} maps a high-precision value into discrete levels. 
We study integer uniform quantization~\cite{jacob2018quantization} (specifically INT8) for better hardware support and efficiency. The quantization process can be expressed as:
\begin{equation}
    \label{eqn:quantizer}
    \bar{\mathbf{X}}^{\text{INT8}} =  \lceil\frac{\mathbf{X^{\text{FP16}}}}{\Delta}\rfloor, \quad \Delta=\frac{\max(|\mathbf{X}|)}{2^{N-1}-1},
\end{equation}
where $\mathbf{X}$ is the floating-point tensor, $\bar{\mathbf{X}}$ is the quantized counterpart, $\Delta$ is the quantization step size, $\lceil\cdot\rfloor$ is the rounding function, and $N$ is the number of bits (8 in our case). Here we assume the tensor is \emph{symmetric} at 0 for simplicity; the discussion is similar for asymmetric cases (\eg, after ReLU) by adding a zero-point~\cite{jacob2018quantization}.

Such quantizer uses the maximum absolute value to calculate $\Delta$ so that it preserves the outliers in activation, which are found to be important for accuracy~\cite{dettmers2022llmint8}. We can calculate $\Delta$ offline with the activations of some calibration samples, what we call \textbf{static quantization}. We can also use the runtime statistics of activations to get $\Delta$, what we call \textbf{dynamic quantization}.
\begin{figure}[t]
    \centering
    \includegraphics[width=0.37\textwidth]{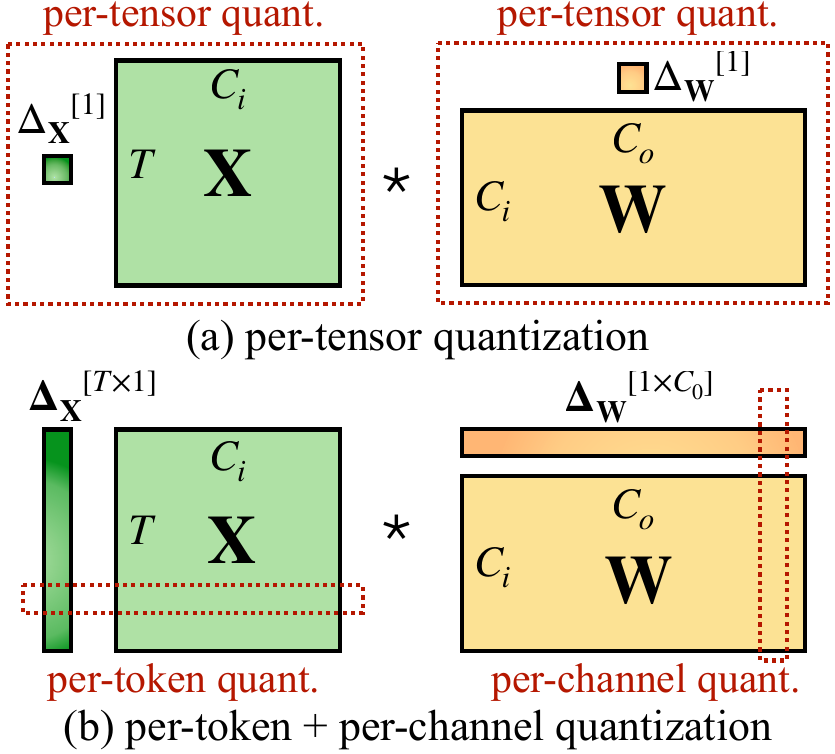}
    \caption{
       Definition of per-tensor, per-token, and per-channel quantization. Per-tensor quantization is the most efficient to implement. For vector-wise quantization to efficiently utilize the INT8 \texttt{GEMM} kernels, we can only use scaling factors from the outer dimensions (\ie, token dimension $T$ and out channel dimension $C_o$) but not inner dimension (\ie, in channel dimension $C_i$). 
    }
    \label{fig:quantize_notation}
    \vspace{-10pt}
   \end{figure}

As shown in Figure~\ref{fig:quantize_notation}, quantization has different granularity levels. The \textbf{per-tensor} quantization uses a single step size for the entire matrix. We can further enable finer-grained quantization by using different quantization step sizes for activations associated with each token (\textbf{per-token} quantization) or each output channel of weights (\textbf{per-channel} quantization). A coarse-grained version of per-channel quantization is to use different quantization steps for different channel groups, called \textbf{group-wise} quantization~\cite{shen2020q,zeroquant}.

For a linear layer in Transformers~\cite{vaswani2017attention}
$\mathbf{Y}=\mathbf{X}\cdot \mathbf{W}, \mathbf{Y}\in\mathbb{R}^{T\times C_o}, \mathbf{X}\in\mathbb{R}^{T\times C_i}, \mathbf{W}\in\mathbb{R}^{C_i\times C_o}$,
where  $T$ is the number of tokens, $C_i$ is the input channel, and $C_o$ is the output channel (see Figure~\ref{fig:quantize_notation}, we omit the batch dimension for simplicity), we can reduce the storage by half compared to FP16 by quantizing the weights to INT8. However, to speed up the inference, we need to quantize both weights and activations into INT8 (\ie, W8A8) to utilize the integer kernels (\eg, INT8 \texttt{GEMM}), which are supported by a wide range of hardware (\eg, NVIDIA GPUs, Intel CPUs, Qualcomm DSPs, \etc).
\begin{figure*}[h]
    \centering
    \includegraphics[width=0.95\textwidth]{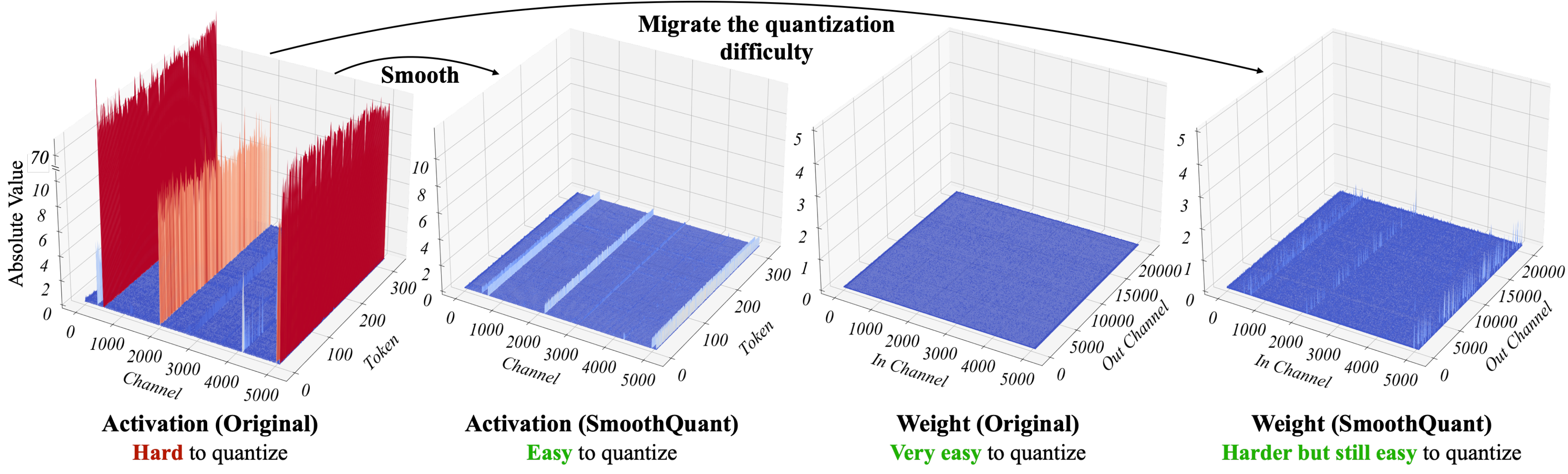}
    \caption{
        Magnitude of the input activations and weights of a linear layer in OPT-13B before and after \method.
        Observations: (1) there are a few channels in the original activation map whose magnitudes are very large (greater than 70); (2) the variance in one activation channel is small; (3) the original weight distribution is flat and uniform. \method migrates the outlier channels from activation to weight. In the end, the outliers in the activation are greatly smoothed while the weight is still pretty smooth and flat.
    }
    \label{fig:aw-dist}
    \vspace{-10pt}
\end{figure*}

\section{Review of Quantization Difficulty}
LLMs are notoriously difficult to quantize due to the outliers in the activations~\citep{dettmers2022llmint8,outlier_suppression,bondarenko2021understanding}.
We first review the difficulties of activation quantization and look for a pattern amongst outliers.
We visualize the input activations and the weights of a linear layer that has a large quantization error in Figure~\ref{fig:aw-dist} (left). We can find several patterns that motivate our method:

\textbf{1. Activations are harder to quantize than weights. } The weight distribution is quite uniform and flat, which is easy to quantize. Previous work has shown that quantizing the weights of LLMs with INT8 or even with INT4 does not degrade accuracy~\cite{dettmers2022llmint8,zeroquant,zeng2022glm}, which echoes our observation.

\textbf{2. Outliers make activation quantization difficult.} The scale of outliers in activations is $\sim 100\times$ larger than most of the activation values.
In the case of per-tensor quantization (\eqn{eqn:quantizer}), the large outliers dominate the maximum magnitude measurement, leading to low \emph{effective quantization bits/levels} (Figure~\ref{fig:intuition}) for non-outlier channels:
suppose the maximum magnitude of channel $i$ is $m_i$, and the maximum value of the whole matrix is $m$, the effective quantization levels of channel $i$ is $2^8 \cdot m_i/m$. For non-outlier channels, the effective quantization levels would be very small (2-3), leading to large quantization errors.

\textbf{3. Outliers persist in fixed channels. } Outliers appear in a small fraction of the \emph{channels}. If one channel has an outlier, it persistently appears in all tokens (Figure \ref{fig:aw-dist}, red).
The variance amongst the channels for a given token is large (the activations in some channels are very large, but most are small), but the variance between the magnitudes of a given channel across tokens is small (outlier channels are consistently large).
\definecolor{mylightgray}{rgb}{0.7, 0.7,0.7}
\newcommand{\textgray}[1]{\textcolor{mylightgray}{{#1}}}
\begin{table}[t]
    \setlength{\tabcolsep}{3pt}
    \small
    \centering
    \caption{Among different activation quantization schemes, only per-channel quantization~\cite{bondarenko2021understanding} preserves the accuracy, but it is \emph{not} compatible (marked in \textgray{gray}) with INT8 \texttt{GEMM} kernels. 
    We report the average accuracy on WinoGrande, HellaSwag, PIQA, and LAMBADA.
    }
    \label{tab:per-feature}
    \begin{tabular}{lccccc}
        \toprule
        Model size (OPT-)                  & {6.7B}            & {13B}             & {30B}             & {66B}             & {175B}            \\
        \midrule
        FP16                        & 64.9\%            & 65.6\%            & 67.9\%            & 69.5\%            & 71.6\%            \\
        \midrule
        INT8 per-tensor             & 39.9\%            & 33.0\%            & 32.8\%            & 33.1\%            & 32.3\%            \\
        INT8 per-token              & 42.5\%            & 33.0\%            & 33.1\%            & 32.9\%            & 31.7\%            \\
        \textgray{INT8 per-channel} & \textgray{64.8\%} & \textgray{65.6\%} & \textgray{68.0\%} & \textgray{69.4\%} & \textgray{71.4\%} \\
        \bottomrule
    \end{tabular}
\end{table}

Due to the persistence of outliers and the small variance inside each channel, if we could perform \emph{per-channel} quantization~\cite{bondarenko2021understanding} of the activation (\ie, using a different quantization step for each channel), the quantization error would be much smaller compared to \emph{per-tensor} quantization, while \emph{per-token} quantization helps little.
In Table~\ref{tab:per-feature}, we verify the assumption that \textit{simulated} per-channel activation quantization successfully bridges the accuracy with the FP16 baseline, which echos the findings of \citeauthor{bondarenko2021understanding}.

However, per-channel activation quantization does not map well to hardware-accelerated \texttt{GEMM} kernels, that rely on a sequence of operations executed at a high throughput (\eg, Tensor Core MMAs) and do not tolerate the insertion of instructions with a lower throughput (\eg, conversions or CUDA Core FMAs) in that sequence. In those kernels, scaling can only be performed along the outer dimensions of the matrix multiplication (\ie, token dimension of activations $T$, output channel dimension of weights $C_o$, see Figure~\ref{fig:quantize_notation}), which can be applied after the matrix multiplication finishes:
\begin{equation}
    \mathbf{Y} = \text{diag}(\mathbf{\Delta}_{\mathbf{X}}^{\text{FP16}}) \cdot (\mathbf{\bar{X}}^{\text{INT8}}\cdot \mathbf{\bar{W}}^{\text{INT8}}) \cdot \text{diag}(\mathbf{\Delta}_{\mathbf{W}}^{\text{FP16}})
\end{equation}
Therefore, previous works all use per-token activation quantization for linear layers~\cite{dettmers2022llmint8, zeroquant}, although they cannot address the difficulty of activation quantization (only slightly better than per-tensor).

\section{\method}
Instead of per-channel activation quantization (which is infeasible), we propose to ``smooth'' the input activation by dividing it by a per-channel smoothing factor $\mathbf{s}\in\mathbb{R}^{C_i}$. To keep the mathematical equivalence of a linear layer, we scale the weights accordingly in the reversed direction:
\begin{equation}
    \mathbf{Y} = (\mathbf{X}\text{diag}(\mathbf{s})^{-1}) \cdot(\text{diag}(\mathbf{s})\mathbf{W}) = \hat{\mathbf{X}} \hat{\mathbf{W}}
\end{equation}
Considering input $\mathbf{X}$ is usually produced from previous linear operations (\eg, linear layers, layer norms, \etc), we can easily fuse the smoothing factor into previous layers' parameters \textit{offline}, which doe not incur kernel call overhead from an extra scaling. For some other cases, when the input is from a residual add, we can add an extra scaling to the residual branch similar to~\citet{outlier_suppression}.

\myparagraph{Migrate the quantization difficulty from activations to weights. }
We aim to choose a per-channel smoothing factor $\mathbf{s}$ such that $\hat{\mathbf{X}}=\mathbf{X}\text{diag}(\mathbf{s})^{-1}$ is easy to quantize.
To reduce the quantization error, we should \emph{increase the effective quantization bits}
for all the channels. The total effective quantization bits would be largest when all the channels have the same maximum magnitude. Therefore, a straight-forward choice is $\mathbf{s}_j = \max(|\mathbf{X}_j|), j=1,2,..., C_i$, where $j$ corresponds to $j$-th input channel. This choice ensures that after the division, all the activation channels will have the same maximum value, which is easy to quantize.
Note that the range of activations is dynamic; it varies for different input samples. Here, we estimate the scale of activations channels using calibration samples from the pre-training dataset~\cite{jacob2018quantization}.
However, this formula pushes \emph{all} the quantization difficulties to the weights. We find that, in this case, the quantization errors would be large for the weights (outlier channels are migrated to weights now), leading to a large accuracy degradation (see \fig{fig:alpha}).
On the other hand, we can also push all the quantization difficulty from weights to activations by choosing $\mathbf{s}_j = 1/\max(|\mathbf{W}_j|)$. Similarly, the model performance is bad due to the activation quantization errors.
Therefore, we need to \emph{split} the quantization difficulty between weights and activations so that they are both easy to quantize.
\begin{figure}
    \centering
    \includegraphics[width=0.95\linewidth]{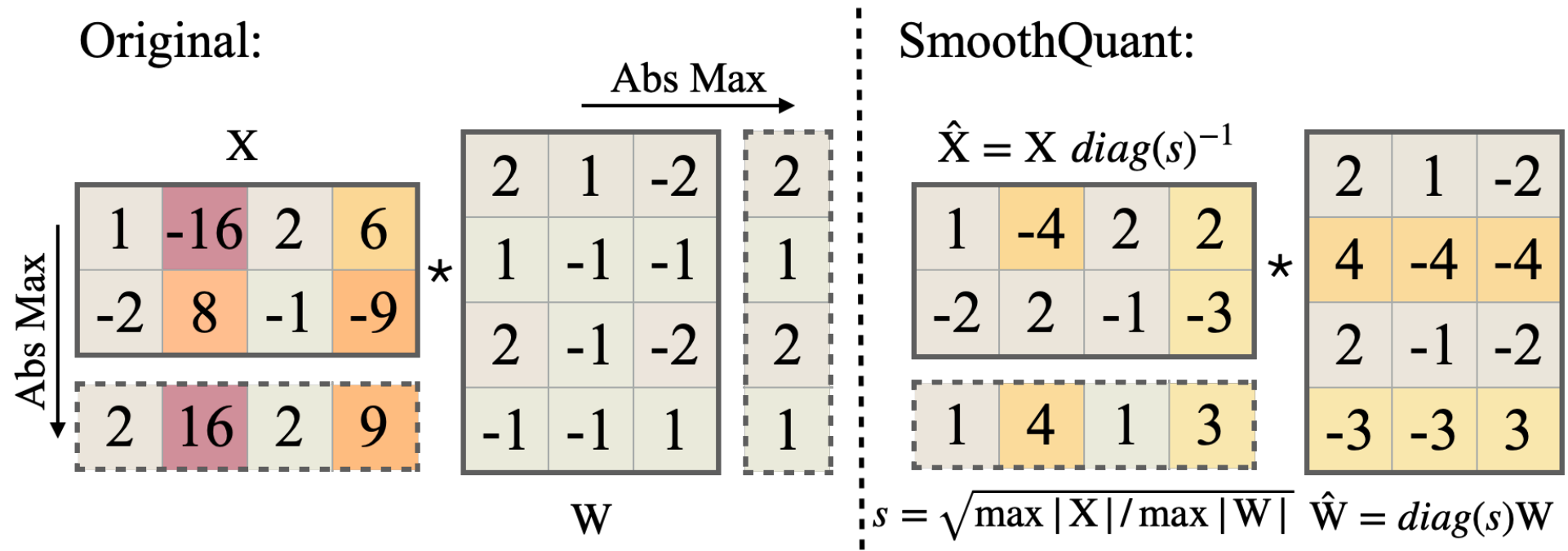}
    \caption{Main idea of \method when $\alpha$ is $0.5$. The smoothing factor $s$ is obtained on calibration samples and the entire transformation is performed offline. At runtime, the activations are smooth without scaling.}
    \label{fig:transform}
    \vspace{-10pt}
\end{figure}

Here we introduce a hyper-parameter, migration strength $\alpha$, to control how much difficulty we want to migrate from activation to weights, using the following equation:
\begin{equation} \label{eq:s_sqrt}
    \mathbf{s}_j =  \max(|\mathbf{X}_j|)^{\alpha} / \max(|\mathbf{W}_j|) ^{1-\alpha}
\end{equation}
We find that for most of the models, \eg, all OPT~\cite{opt} and BLOOM~\cite{scao2022bloom} models, $\alpha=0.5$ is a well-balanced point to evenly split the quantization difficulty, especially when we are using the same quantizer for weights and activations (\eg, per-tensor, static quantization). The formula ensures that the weights and activations at the corresponding channel share a similar maximum value, thus sharing the same quantization difficulty.
\fig{fig:transform} illustrates the smoothing transformation when we take $\alpha=0.5$.
For some other models where activation outliers are more significant (\eg, GLM-130B~\cite{zeng2022glm} has $\sim$30\% outliers, which are more difficult for activation quantization), we can choose a larger $\alpha$ to migrate more quantization difficulty to weights (like 0.75).

\begin{figure}[t]
    \centering
    \includegraphics[width=0.9\linewidth]{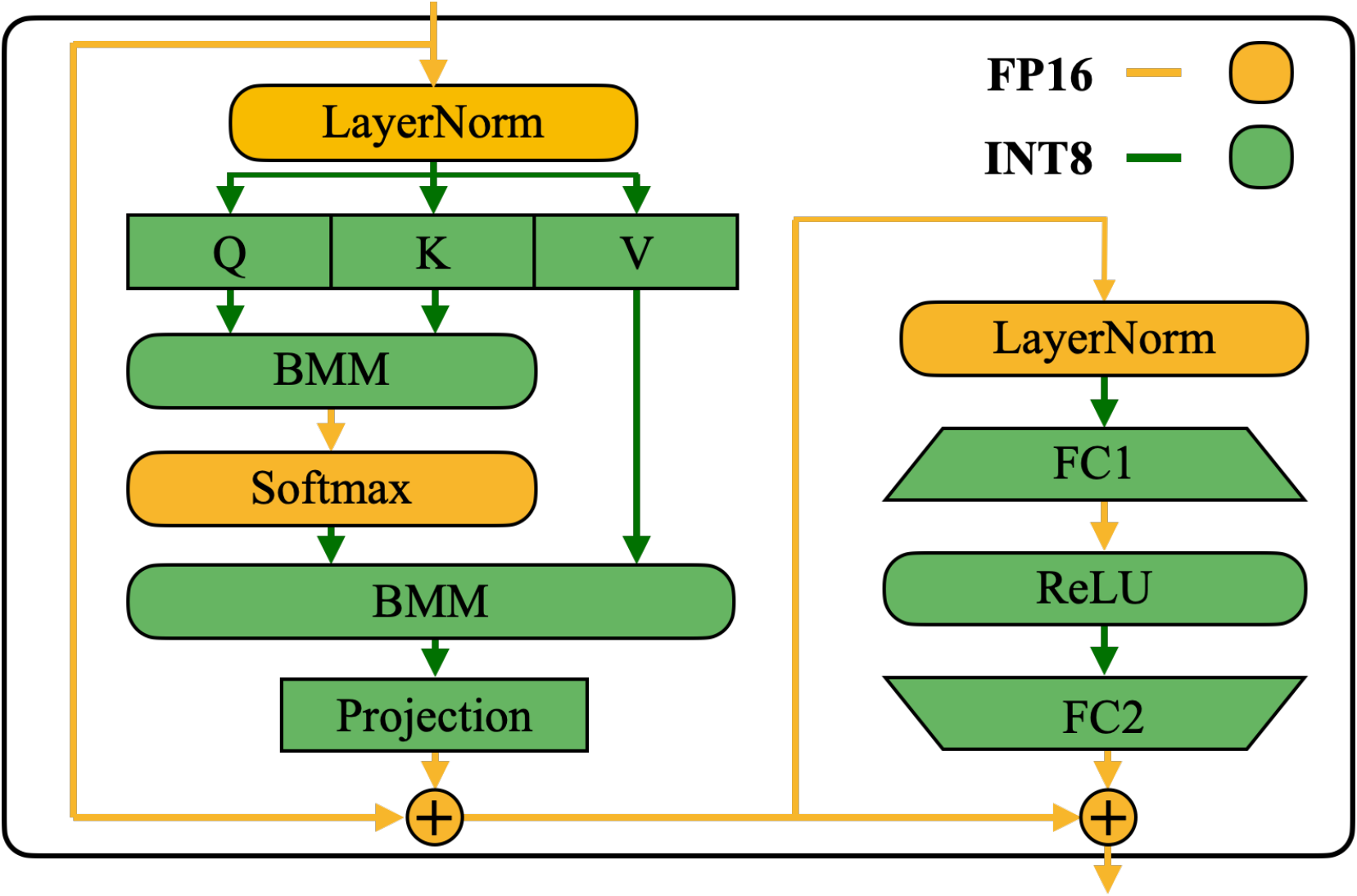}
    \caption{\method's precision mapping for a Transformer block. All compute-intensive operators like linear layers and batched matmul (\texttt{BMM}s) use INT8 arithmetic.
    }
    \label{fig:quantization-flow}
\end{figure}

\myparagraph{Applying \method to Transformer blocks. } Linear layers take up most of the parameters and computation of LLM models. By default, we perform scale smoothing for the input activations of self-attention and feed-forward layers and quantize all linear layers with W8A8. We also quantize \texttt{BMM} operators in the attention computation.
We design a quantization flow for transformer blocks in Figure~\ref{fig:quantization-flow}. We quantize the inputs and weights of compute-heavy operators like linear layers and \texttt{BMM} in attention layers with INT8, while keeping the activation as FP16 for other lightweight element-wise operations like ReLU, Softmax, and LayerNorm. Such a design helps us to balance accuracy and inference efficiency.

\section{Experiments}
\subsection{Setups}
\myparagraph{Baselines.}
\begin{table}[t]
      \setlength{\tabcolsep}{2.5pt}
      \small
      \centering
        \caption{Quantization setting of the baselines and \method. All weight and activations use INT8 representations unless specified. For \method, the efficiency \textbf{improves} from O1 to O3 (\ie, lower latency).
      }
      \label{tab:settings}
      \begin{tabular}{lll}
            \toprule
            Method              & Weight      & Activation             \\
            \midrule
            W8A8                & per-tensor  & per-tensor dynamic     \\
            ZeroQuant           & group-wise & per-token dynamic      \\
            \texttt{LLM.int8()}          & per-channel & per-token dynamic+FP16 \\
            Outlier Suppression & per-tensor  & per-tensor static      \\
            \midrule
            \method-O1          & per-tensor  & per-token dynamic      \\
            \method-O2          & per-tensor  & per-tensor dynamic     \\
            \method-O3          & per-tensor  & per-tensor static      \\
            \bottomrule
      \end{tabular}
\end{table}

We compare with four baselines in the INT8 post-training quantization setting, i.e., without re-training of the model parameters: W8A8 naive quantization, ZeroQuant~\cite{zeroquant},  \texttt{LLM.int8()}~\cite{dettmers2022llmint8}, and Outlier Suppression~\cite{outlier_suppression}. Since \method is orthogonal to the quantization schemes, we provide gradually aggressive and efficient quantization levels from O1 to O3. The detailed quantization schemes of the baselines and \method are shown in \tbl{tab:settings}.
\renewcommand \arraystretch{0.93}
\begin{table*}[t]
    \setlength{\tabcolsep}{2.5pt}
    \small
    \centering
    \caption{\method maintains the accuracy of OPT-175B model after INT8 quantization, even with the most aggressive and most efficient O3 setting (Table \ref{tab:settings}). We extensively benchmark the performance on 7 zero-shot benchmarks (by reporting the average accuracy) and 1 language modeling benchmark (perplexity). {*}For ZeroQuant, we also tried leaving the input activation of self-attention in FP16 and quantizing the rest to INT8, which is their solution to the GPT-NeoX-20B. But this does not solve the accuracy degradation of OPT-175B.}
    \label{tab:175b_acc}
    \begin{tabular}{lccccccccc}
        \toprule
        \emph{OPT-175B}     & {LAMBADA} & {HellaSwag} & {PIQA} & {WinoGrande} & {OpenBookQA} & {RTE}  & {COPA} & \textbf{Average$\uparrow$} & \textbf{WikiText$\downarrow$} \\
        \midrule
        FP16                & 74.7\%    & 59.3\%      & 79.7\% & 72.6\%       & 34.0\%       & 59.9\% & 88.0\% & 66.9\%                     & 10.99                         \\ \midrule
        W8A8                & 0.0\%     & 25.6\%      & 53.4\% & 50.3\%       & 14.0\%       & 49.5\% & 56.0\% & 35.5\%                     & 93080                         \\
        ZeroQuant           & 0.0\%{*}  & 26.0\%      & 51.7\% & 49.3\%       & 17.8\%       & 50.9\% & 55.0\% & 35.8\%                     & 84648                         \\
        \texttt{LLM.int8()}          & 74.7\%    & 59.2\%      & 79.7\% & 72.1\%       & 34.2\%       & 60.3\% & 87.0\% & 66.7\%                     & 11.10                         \\
        Outlier Suppression & 0.00\%    & 25.8\%      & 52.5\% & 48.6\%       & 16.6\%       & 53.4\% & 55.0\% & 36.0\%                     & 96151                         \\ \midrule
        \method-O1 & 74.7\% & 59.2\% & 79.7\% & 71.2\% & 33.4\% & 58.1\% & 89.0\% & 66.5\% & 11.11  \\
        \method-O2 & 75.0\% & 59.0\% & 79.2\% & 71.2\% & 33.0\% & 59.6\% & 88.0\% & 66.4\% & 11.14  \\
        \method-O3 & 74.6\% & 58.9\% & 79.7\% & 71.2\% & 33.4\% & 59.9\% & 90.0\% & 66.8\% & 11.17 \\
        \bottomrule
    \end{tabular}
\end{table*}

\definecolor{mylightgray}{rgb}{0.8, 0.8,0.8}
\newcommand{\hidegray}[1]{{{#1}}}
\renewcommand \arraystretch{0.95}
\begin{table}
    \setlength{\tabcolsep}{1pt}
    \small
    \centering
    \caption{\method works for different LLMs. We can quantize the 3 largest, openly available LLM models into INT8 without degrading the accuracy. For OPT-175B and BLOOM-176B, we show the average accuracy on WinoGrande, HellaSwag, PIQA, and LAMBADA. For GLM-130B we show the average accuracy on LAMBADA, MMLU, MNLI, and QNLI. {*}Accuracy is not column-wise comparable due to different datasets. }
    \label{tab:different_models}
    \begin{tabular}{lccc}
        \toprule
        Method                & OPT-175B          & BLOOM-176B        & GLM-130B{*}       \\
        \midrule
        FP16                  & 71.6\%            & 68.2\%            & 73.8\%            \\
        \midrule
        W8A8                  & 32.3\%            & 64.2\%            & 26.9\%            \\

        ZeroQuant             & 31.7\%            & 67.4\%            & 26.7\%            \\

        \hidegray{\texttt{LLM.int8()}} & \hidegray{71.4\%} & \hidegray{68.0\%} & \hidegray{73.8\%} \\

        Outlier Suppression   & 31.7\%            & 54.1\%            & 63.5\%            \\
        \midrule
        \method-O1            &\textbf{71.2}\%            & 68.3\%            & \textbf{73.7\%}   \\

        \method-O2            & 71.1\%   & \textbf{68.4}\%   & 72.5\%            \\

        \method-O3            & 71.1\%            & 67.4\%            & 72.8\%            \\
        \bottomrule
    \end{tabular}
\end{table}

\myparagraph{Models and datasets.}
We choose three families of LLMs to evaluate \method: OPT~\cite{opt}, BLOOM~\cite{scao2022bloom}, and GLM-130B~\cite{zeng2022glm}.
We use seven zero-shot evaluation tasks: LAMBADA~\cite{paperno-etal-2016-lambada}, HellaSwag~\cite{hellaswag}, PIQA~\cite{Bisk2020piqa}, WinoGrande~\cite{sakaguchi2019winogrande}, OpenBookQA~\cite{OpenBookQA2018}, RTE~\cite{glue}, COPA~\cite{copa}, and one language modeling dataset WikiText~\cite{merity2016pointer} to evaluate the OPT and BLOOM models. We use MMLU~\cite{mmlu}, MNLI~\cite{mnli}, QNLI~\cite{glue} and LAMBADA to evaluate the GLM-130B model because some of the aforementioned benchmarks appear in the training set of GLM-130B.
We use {\texttt{lm-eval-harness}}\footnote{\url{https://github.com/EleutherAI/lm-evaluation-harness}} to evaluate OPT and BLOOM models, and GLM-130B's official repo\footnote{\url{https://github.com/THUDM/GLM-130B}} for its own evaluation. Finally, we scale up our method to MT-NLG 530B~\cite{smith2022using} and for the first time enabling the serving of a >500B model within a single node.
Note that we focus on the \emph{relative} performance change before and after quantization but not the absolute value.

\myparagraph{Activation smoothing.} The migration strength $\alpha=0.5$ is a general sweet spot for all the OPT and BLOOM models, and $\alpha=0.75$ for GLM-130B since its activations are more difficult to quantize~\cite{zeng2022glm}.
We get a suitable $\alpha$ by running a quick grid search on a subset of the Pile~\cite{gao2020pile} validation set.
To get the statistics of activations, we calibrate the smoothing factors and the static quantization step sizes \emph{once} with 512 random sentences from the pre-training dataset Pile, and apply the same smoothed and quantized model for all downstream tasks. In this way, we can benchmark the generality and zero-shot performance of the quantized LLMs.

\myparagraph{Implementation.}
We implement \method with two backends: (1) PyTorch Huggingface\footnote{\url{https://github.com/huggingface/transformers}} for the proof of concept, and (2) FasterTransformer\footnote{\url{https://github.com/NVIDIA/FasterTransformer}}, as an example of a high-performance framework used in production environments. In both PyTorch Huggingface and FasterTransformer frameworks, we implement INT8 linear modules and the batched matrix multiplication (BMM) function with CUTLASS INT8 \texttt{GEMM} kernels. We simply replace the original floating point (FP16) linear modules and the \texttt{bmm} function with our INT8 kernels as the INT8 model.

\subsection{Accurate Quantization}
\label{sec:accurate-quantization}
\myparagraph{Results of OPT-175B.}
\method can handle the quantization of very large LLMs, whose activations are more difficult to quantize.
We study quantization on OPT-175B.
As shown in \tbl{tab:175b_acc}, \method can match the FP16 accuracy on all evaluation datasets with all quantization schemes. \texttt{LLM.int8()} can match the floating point accuracy because they use floating-point values to represent outliers, which leads to a large latency overhead (Table~\ref{tab:quant_schemes_latency}). The W8A8, ZeroQuant, and Outlier Suppression baselines produce nearly random results, indicating that naively quantizing the activation of LLMs will destroy the performance.
\myparagraph{Results of different LLMs.}
\method can be applied to various LLM designs.
In \tbl{tab:different_models}, we show \method can quantize all existing open LLMs beyond 100B parameters.
Compared with the OPT-175B model, the BLOOM-176B model is easier to quantize: none of the baselines completely destroys the model; even the naive W8A8 per-tensor dynamic quantization only degrades the accuracy by 4\%. The O1 and O2 levels of \method successfully maintain the floating point accuracy, while the O3 level (per-tensor static) degrades the average accuracy by 0.8\%, which we attribute to the discrepancy between the statically collected statistics and the real evaluation samples' activation statistics.
On the contrary, the GLM-130B model is more difficult to quantize (which echos~\citeauthor{zeng2022glm}). Nonetheless, \method-O1 can match the FP16 accuracy, while \method-O3 only degrades the accuracy by 1\%, which significantly outperforms the baselines. Note that we clip the top 2\% tokens when calibrating the static quantization step sizes for GLM-130B following~\citet{outlier_suppression}.
Note that different model/training designs have different quantization difficulties, which we hope will inspire future research.
\begin{figure}
    \centering
    \includegraphics[width=0.93\linewidth]{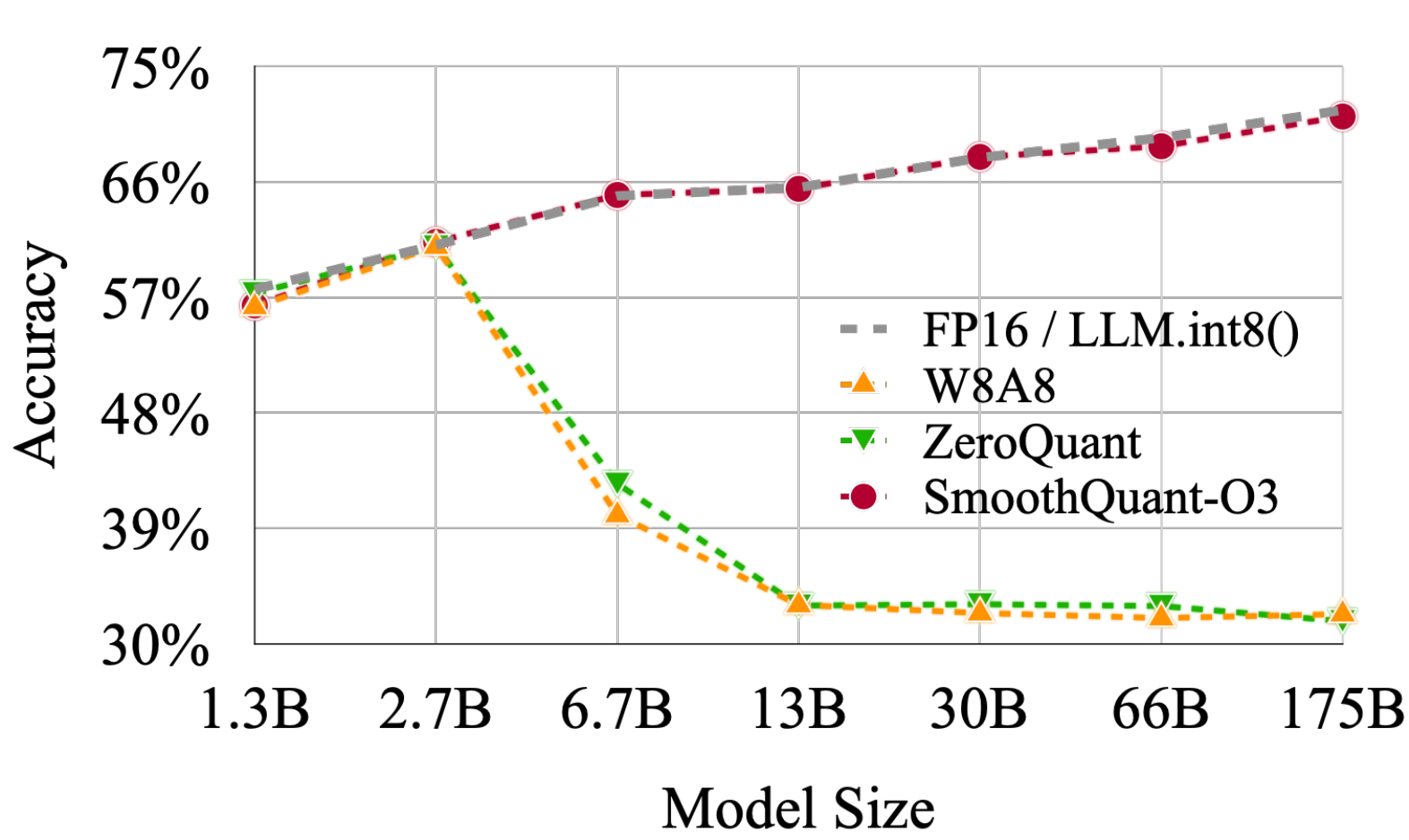}
    \caption{\method-O3 (the most efficient setting, defined in Table \ref{tab:settings}) preserves the accuracy of OPT models across different scales when quantized to INT8. LLM.int8() requires  mixed precision and suffers from slowing down.
    }
    \label{fig:model_size}
\end{figure}

\myparagraph{Results on LLMs of different sizes.} \method works not only for the very large LLMs beyond 100B parameters, but it also works consistently for smaller LLMs.
In \fig{fig:model_size}, we show that \method can work on all scales of OPT models, matching the FP16 accuracy with INT8 quantization.

\myparagraph{Results on Instruction-Tuned LLM}
\begin{table}
\setlength{\tabcolsep}{2.5pt}
\small
\centering
\caption{SmoothQuant's performance on the OPT-IML model.
}
\label{tab:opt-iml}
\begin{tabular}{lcc}
\toprule
OPT-IML-30B        & LAMBADA $\uparrow$ & WikiText $\downarrow$  \\
\midrule
FP16               & 69.12\%                     & 14.26                           \\
\midrule
W8A8               & 4.21\%                      & 576.53                          \\
ZeroQuant          & 5.12\%                      & 455.12                          \\
LLM.int8()         & 69.14\%                     & 14.27                           \\
Outlier Suppression & 0.00\%                      & 9485.62                         \\
\midrule
SmoothQuant-O3     &\textbf{ 69.77\%}                     & \textbf{14.37}\\
\bottomrule
\end{tabular}
\end{table}

Shown in Table~\ref{tab:opt-iml}, SmoothQuant also works on instruction-tuned LLMs. We test SmoothQuant on the OPT-IML-30B model using the WikiText-2 and LAMBADA datasets. Our results show that SmoothQuant successfully preserves model accuracy with W8A8 quantization, whereas the baselines fail to do so. SmoothQuant is a general method designed to balance the quantization difficulty for Transformer models. As the architecture of instruction-tuned LLMs is not fundamentally different from vanilla LLMs, and their pre-training processes are very similar, SmoothQuant is applicable to instruction-tuned LLMs as well.

\myparagraph{Results on LLaMA models.}
\begin{table}
\setlength{\tabcolsep}{8pt}
\small
\centering
\caption{SmoothQuant can enable lossless W8A8 quantization for LLaMA models~\cite{touvron2023llama}. Results are perplexities on the WikiText-2 dataset with a sequence length of 512. We used per-token activation quantization and $\alpha$=0.8 for SmoothQuant. 
}
\label{tab:llama}
\begin{tabular}{lcccc}
\toprule
Wiki PPL$\downarrow$  & 7B & 13B & 30B & 65B  \\
\midrule
FP16  & 11.51 & 10.05 & 7.53 & 6.17    \\
W8A8 SmoothQuant & 11.56 & 10.08 & 7.56 & 6.20\\
\bottomrule
\end{tabular}
\end{table}

LLaMA models are new open languange models with superior performance~\cite{touvron2023llama}. Through initial experiments, we find LLaMA models generally have less severe activation outlier issues compared to models like OPT and BLOOM. Nonetheless, SmoothQuant still works quite well for LLaMA models. We provide some initial results of LLaMA W8A8 quantization in Table~\ref{tab:llama}. SmoothQuant enables W8A8 quantization at a negligible performance degradation. 

\myparagraph{Results on Llama-2, Falcon, Mistral, and Mixtral models.}
\begin{table}
\setlength{\tabcolsep}{8pt}
\small
\centering
\caption{SmoothQuant can enable lossless W8A8 quantization for Llama-2~\cite{touvron2023llama2}, Falcon~\cite{almazrouei2023falcon}, Mistral~\cite{jiang2023mistral}, and Mixtral~\cite{jiang2024mixtral} models. Results are perplexities on the WikiText-2 dataset with a sequence length of 2048. We used per-token activation quantization and per-channel weight quantization for SmoothQuant. 
}
\label{tab:more_models}
    \begin{tabular}{llcc}
        \toprule
        Model        & Method    & PPL   & $\alpha$ \\
        \midrule
        \multirow{2}{*}{Llama-2-7B}   & FP16      & 5.474 &       \\
                                       & W8A8 SQ   & 5.515 & 0.85  \\
        \midrule
        \multirow{2}{*}{Llama-2-13B}  & FP16      & 4.950 &       \\
                                       & W8A8 SQ   & 4.929 & 0.85  \\
        \midrule
        \multirow{2}{*}{Llama-2-70B}  & FP16      & 3.320 &       \\
                                       & W8A8 SQ   & 3.359 & 0.9   \\
        \midrule
        \multirow{2}{*}{Falcon-7B}    & FP16      & 6.590 &       \\
                                       & W8A8 SQ   & 6.629 & 0.6   \\
        \midrule
        \multirow{2}{*}{Falcon-40B}   & FP16      & 5.228 &       \\
                                       & W8A8 SQ   & 5.255 & 0.7   \\
        \midrule
        \multirow{2}{*}{Mistral-7B}   & FP16      & 5.253 &       \\
                                       & W8A8 SQ   & 5.277 & 0.8   \\
        \midrule
        \multirow{2}{*}{Mixtral-8x7B} & FP16      & 3.842 &       \\
                                       & W8A8 SQ   & 3.893 & 0.8   \\
        \bottomrule
    \end{tabular}
\end{table}

We apply \method on several more recent LLMs using diverse architectures, such as Llama-2~\cite{touvron2023llama2}, Falcon~\cite{almazrouei2023falcon}, Mistral~\cite{jiang2023mistral}, and Mixtral~\cite{jiang2024mixtral}—notably, the Mixtral model is a Mixture of Experts (MoE) model. The results, detailed in Table~\ref{tab:more_models}, demonstrate that SmoothQuant enables W8A8 quantization while maintaining performance with minimal loss across these varied architectures.

\subsection{Speedup and Memory Saving}
In this section, we show the measured speedup and memory saving of \method-O3 integrated into PyTorch and FasterTransformer.
\myparagraph{Context-stage: PyTorch Implementation.}
We measure the end-to-end latency of generating all hidden states for a batch of 4 sentences in one pass, \ie, the context stage latency. We record the (aggregated) peak GPU memory usage in this process. We only compare \method with \texttt{LLM.int8()} because it is the only existing quantization method that can preserve LLM accuracy at all scales.
Due to the lack of support for model parallelism in Huggingface, we only measure \method's performance on a single GPU for the PyTorch implementation, so we choose OPT-6.7B, OPT-13B, and OPT-30B for evaluation. In the FasterTransformer library, \method can seamlessly work with Tensor Parallelism~\cite{megatron_lm} algorithm, so we test \method on OPT-13B, OPT-30B, OPT-66B, and OPT-175B for both single and multi-GPU benchmarks. All our experiments are conducted on NVIDIA A100 80GB GPU servers.

\begin{figure}[t]
    \centering
    \includegraphics[width=\linewidth]{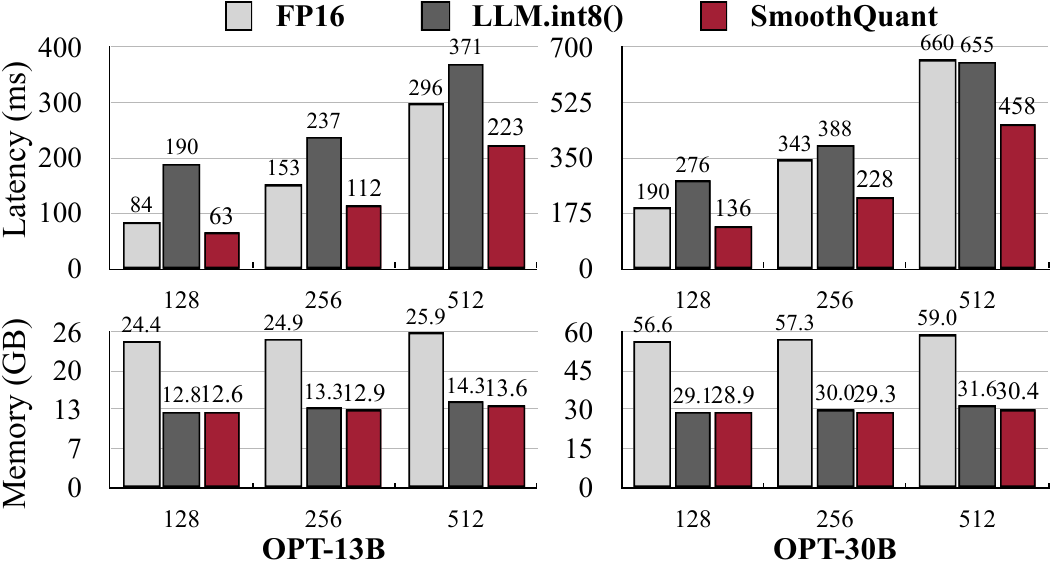}
    \vspace{-10pt}
    \caption{The PyTorch implementation of \method-O3 achieves up to \textbf{1.51$\times$}  speedup and \textbf{1.96$\times$} memory saving for OPT models on a single NVIDIA A100-80GB GPU, while \texttt{LLM.int8()} slows down the inference in most cases.}
    \vspace{-10pt}
    \label{fig:pytorch-results}
\end{figure}

In \fig{fig:pytorch-results}, we show the inference latency and peak memory usage based on the PyTorch implementation. \method is consistently faster than the FP16 baseline, getting a 1.51x speedup on OPT-30B when the sequence length is 256. We also see a trend that the larger the model, the more significant the acceleration. On the other hand, \texttt{LLM.int8()} is almost always slower than the FP16 baseline, which is due to the large overhead of the mixed-precision activation representation. In terms of memory, \method and \texttt{LLM.int8()} can all nearly halve the memory usage of the FP16 model, while \method saves slightly more memory because it uses fully INT8 \texttt{GEMM}s.

\begin{figure*}[t]
    \centering
    \includegraphics[width=\textwidth]{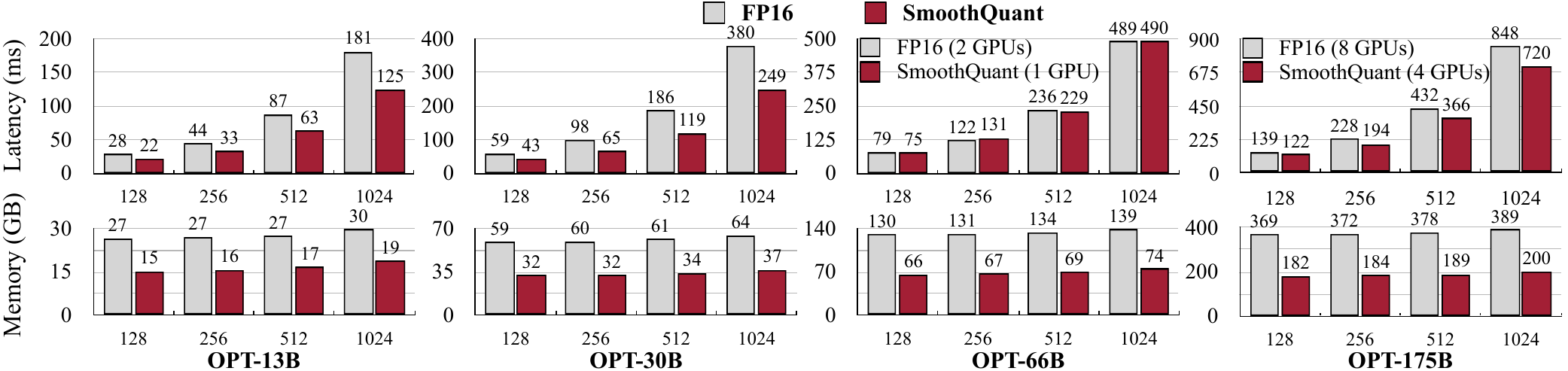}
    \caption{Inference latency (top) and memory usage (bottom) of the FasterTransformer implementation on NVIDIA A100-80GB GPUs. For smaller models, the latency can be significantly reduced with \method-O3 by up to 1.56x compared to FP16.
        For the bigger models (OPT-66B and 175B), we can achieve similar or even faster inference using only \textbf{half} number of GPUs.
        Memory footprint is almost halved compared to FP16.}
    \label{fig:ft-results}
\end{figure*}

\myparagraph{Context-stage: FasterTransformer Implementation.}
As shown in \fig{fig:ft-results}~(top), compared to FasterTransformer's FP16 implementation of OPT, \method-O3 can further reduce the execution latency of OPT-13B and OPT-30B by up to 1.56$\times$ when using a single GPU.
This is challenging since FasterTransformer is already more than 3$\times$ faster compared to the PyTorch implementation for OPT-30B.
Remarkably, for bigger models that have to be distributed across multiple GPUs, \method achieves similar or even better latency using only \emph{half} the number of GPUs (1 GPU instead of 2 for OPT-66B, 4 GPUs instead of 8 for OPT-175B). This could greatly lower the cost of serving LLMs. The amount of memory needed when using \method-O3 in FasterTransformer is reduced by a factor of almost 2$\times$, as shown on \fig{fig:ft-results}~(bottom).

\myparagraph{Decoding-stage.}
In Table~\ref{tab:generation-stage}, we show \method can significantly accelerate the autoregressive decoding stage of LLMs. \method constantly reduces the per-token decoding latency compared to FP16 (up to 1.42x speedup). Additionally, \method halves the memory footprints for LLM inference, enabling the deployment of LLMs at a significantly lower cost.
\begin{table}
\setlength{\tabcolsep}{2.5pt}
\small
\centering
\caption{\method’s performance in the decoding stage.
}
\label{tab:generation-stage}
\small
\begin{tabular}{cccccccc}
\toprule
\multirow{2}{*}{BS} & \multirow{2}{*}{SeqLen} & \multicolumn{3}{c}{Latency (ms)}                       & \multicolumn{3}{c}{Memory (GB)}         \\
\cmidrule(lr){3-5} \cmidrule(lr){6-8}
           &                & FP16                            & Ours & Speedup ($\uparrow$) & FP16        & Ours & Saving ($\uparrow$) \\
           \midrule
\multicolumn{8}{c}{OPT-30B (1 GPU)}        \\
1          & 512            & 422                             & 314         & 1.35$\times$    & 57          & 30          & 1.91$\times$   \\
1          & 1024           & 559                             & 440         & 1.27$\times$    & 58          & 31          & 1.87$\times$   \\
16         & 512            & 2488                            & 1753        & 1.42$\times$    & 69          & 44          & 1.59$\times$   \\
16         & 1024           & OOM                             & 3947        & -       & OOM         & 61          & -      \\
\midrule
\multicolumn{8}{c}{OPT-175B (8 GPUs)}        \\
1          & 512            & 426                             & 359         & 1.19$\times$    & 44          & 23          & 1.87$\times$   \\
1          & 1024           & 571                             & 475         & 1.20$\times$    & 44          & 24          & 1.85$\times$   \\
16         & 512            & 2212                            & 1628        & 1.36$\times$    & 50          & 30          & 1.67$\times$   \\
16         & 1024           & 4133                            & 3231        & 1.28$\times$    & 56          & 37          & 1.52$\times$   \\
\bottomrule
\end{tabular}
\vspace{-1em}
\end{table}

\begin{table}[h]
    \setlength{\tabcolsep}{3pt}
    \small
    \centering
    \caption{\method can quantize MT-NLG 530B to W8A8 with negligible accuracy loss.}
    \begin{tabular}{lccccc}
        \toprule
        & {LAMBADA} & {HellaSwag} & {PIQA} & {WinoGrande} & Average  \\
         \midrule 
        FP16  & 76.6\% & 62.1\% & 81.0\% & 72.9\% & 73.1\%         \\ 
         INT8 & 77.2\% & 60.4\%  & 80.7\% & 74.1\% & 73.1\%  \\
        \bottomrule
    \end{tabular}
    \vspace{-10pt}
    \label{tab:530b_acc}
\end{table}

\begin{table}[h]
    \setlength{\tabcolsep}{7pt}
    \small
    \centering
    \caption{When serving MT-NLG 530B, \method can reduce the memory by half at a similar latency using \emph{half} number of GPUs, which allows serving the 530B model within a single node.}
    \begin{tabular}{llccc}
        \toprule
    {SeqLen} & {Prec.} & {\#GPUs}&  {Latency} & {Memory}  \\
         \midrule 
         128 & FP16 & 16 & 232ms & 1040GB   \\
             & INT8 & 8 & 253ms & 527GB \\ \midrule
         256 &  FP16 & 16 & 451ms & 1054GB  \\
             &  INT8 & 8 & 434ms & 533GB \\ \midrule
         512 &  FP16 & 16 & 838ms & 1068GB  \\
             &  INT8 & 8  & 839ms & 545GB \\ \midrule
        1024 &  FP16 & 16 & 1707ms & 1095GB \\
             &  INT8 & 8  & 1689ms & 570GB \\
        \bottomrule
    \end{tabular}
    \label{tab:530b_latency}
\end{table}

\subsection{Scaling Up: 530B Model Within a Single Node}
We can further scale up \method beyond 500B-level models, enabling efficient and accurate W8A8 quantization of MT-NLG 530B~\cite{smith2022using}. As shown in Table~\ref{tab:530b_acc} and~\ref{tab:530b_latency}, \method enables W8A8 quantization of the 530B model at a negligible accuracy loss. The reduced model size allows us to serve the model using half number of the GPUs (16 to 8) at a similar latency, enabling the serving of a >500B model within a single node (8$\times$A100 80GB GPUs). 

\begin{table}[t]
    \setlength{\tabcolsep}{6pt}
    \small
    \centering
    \caption{GPU Latency (ms) of different quantization schemes. The coarser the quantization scheme (from per-token to per-tensor, dynamic to static, O1 to O3, defined in Table \ref{tab:settings}), the lower the latency. \method achieves lower latency compared to FP16 under all settings, while \texttt{LLM.int8()} is mostly slower.
        The batch size is 4.}
    \label{tab:quant_schemes_latency}
    \begin{tabular}{lcccc}
        \toprule
        \multicolumn{1}{c}{Model} & \multicolumn{2}{c}{OPT-13B} & \multicolumn{2}{c}{OPT-30B}                 \\
        \cmidrule(lr){2-3}\cmidrule(lr){4-5}
        Sequence Length           & 256                         & 512                         & 256   & 512   \\
        \midrule
        FP16                      & 152.6                       & 296.3                       & 343.0 & 659.9 \\
        \texttt{LLM.int8()}                & 237.1                       & 371.5                       & 387.9 & 654.9 \\
        \midrule
        \method-O1                & 124.5                       & 243.3                       & 246.7 & 490.7 \\
        \method-O2                & 120.5                       & 235.1                       & 240.2 & 478.3 \\
        \method-O3                & 112.1                       & 223.1                       & 227.6 & 458.4 \\
        \bottomrule
    \end{tabular}
    \vspace{-5pt}
\end{table}

\subsection{Ablation Study}
\myparagraph{Quantization schemes.}
\tbl{tab:quant_schemes_latency} shows the inference latency of different quantization schemes based on our PyTorch implementation. We can see that the coarser the quantization granularity (from O1 to O3), the lower the latency. And static quantization can significantly accelerate inference compared with dynamic quantization because we no longer need to calculate the quantization step sizes at runtime.
\method is faster than FP16 baseline under all settings, while \texttt{LLM.int8()} is usually slower. We recommend using a coarser scheme if the accuracy permits.

\begin{figure}[t]
    \centering
    \includegraphics[width=\linewidth]{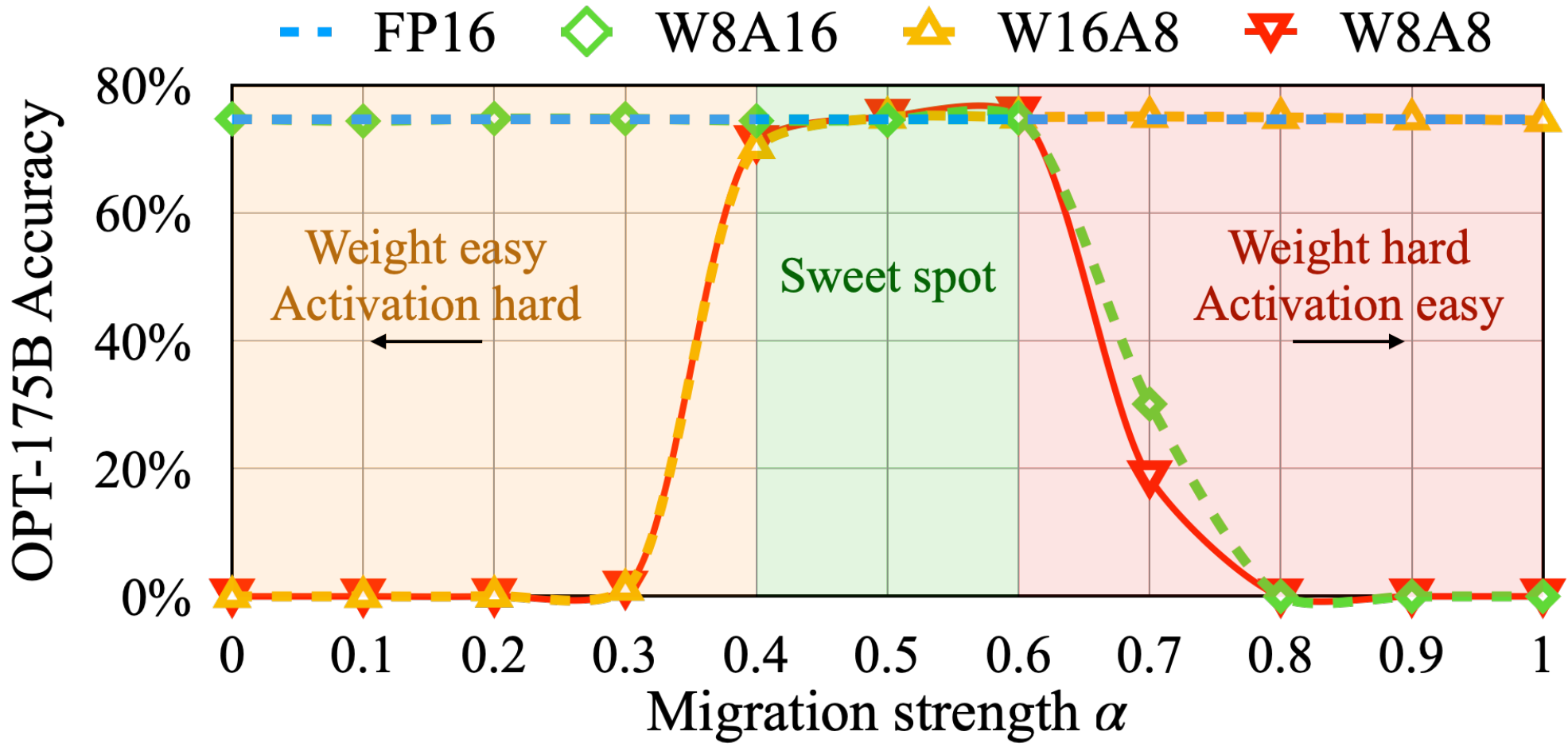}
    \caption{A suitable migration strength $\alpha$ (sweet spot) makes both activations and weights easy to quantize. If the $\alpha$ is too large, weights will be hard to quantize; if too small, activations will be hard to quantize.}
    \vspace{-10pt}
    \label{fig:alpha}
\end{figure}

\myparagraph{Migration strength.}
We need to find a suitable migration strength $\alpha$ (see Equation~\ref{eq:s_sqrt}) to balance the quantization difficulty of weights and activations. We ablate the effect of different $\alpha$'s on OPT-175B with LAMBADA in Figure~\ref{fig:alpha}. When $\alpha$ is too small (<0.4), the activations are hard to quantize; when $\alpha$ is too large (>0.6), the weights will be hard to quantize. Only when we choose $\alpha$ from the sweet spot region (0.4-0.6) can we get small quantization errors for both weights and activations, and maintain the model performance after quantization.

\section{Related Work}
\myparagraph{Large language models (LLMs).}
Pre-trained language models have achieved remarkable performance on various benchmarks by \emph{scaling up}.
GPT-3~\cite{brown2020language} is the first LLM beyond 100B parameters and achieves impressive few-shot/zero-shot learning results. Later works~\cite{rae2021scaling,smith2022using,du2022glam,chowdhery2022palm}  continue to push the frontier of scaling, going beyond 500B parameters.
However, as the language model gets larger, serving such models for inference becomes expensive and challenging. In this work, we show that our proposed method can quantize the three largest, openly available LLMs: OPT-175B~\cite{opt}, BLOOM-176B~\cite{scao2022bloom} and GLM-130B~\cite{zeng2022glm}, and even MT-NLG 530B~\cite{smith2022using} to reduce the memory cost and accelerate inference.

\myparagraph{Model quantization.}
Quantization is an effective method for reducing the model size and accelerating inference. It proves to be effective for various convolutional neural works (CNNs)~\cite{han2016deep,  jacob2018quantization, nagel2019data, wang2019haq, lin2020mcunet} and transformers~\cite{shen2020q, kim2021bert, liu2021post,spatten,bondarenko2021understanding}. Weight equalization~\cite{nagel2019data} and channel splitting~\cite{zhao2019improving} reduce quantization error by suppressing the outliers in weights. However, these techniques cannot address the activation outliers, which are the major quantization bottleneck for LLMs~\cite{dettmers2022llmint8}.

\myparagraph{Quantization of LLMs. }
GPTQ~\cite{frantar2022gptq} applies quantization only to weights but not activations (please find a short discussion in \sectapp{sec:supp_gptq_compare}).
ZeroQuant~\cite{zeroquant} and nuQmm~\cite{park2022nuqmm} use a per-token and group-wise quantization scheme for LLMs, which requires customized CUDA kernels. Their largest evaluated models are 20B and 2.7B, respectively and fail to maintain the performance of LLMs like OPT-175B.
\texttt{LLM.int8()}~\cite{dettmers2022llmint8} uses mixed INT8/FP16 decomposition to address the activation outliers. However, such implementation leads to large latency overhead, which can be even slower than FP16 inference. Outlier Suppression~\cite{outlier_suppression} uses the non-scaling LayerNorm and token-wise clipping to deal with the activation outliers. However, it only succeeds on small language models such as BERT~\cite{bert} and BART~\cite{lewis2019bart} and fails to  maintain the accuracy for LLMs (Table \ref{tab:different_models}).
Our algorithm preserves the performance of LLMs (up to 176B, the largest open-source LLM we can find) with an efficient per-tensor, static quantization scheme without retraining, allowing us to use off-the-shelf INT8 \texttt{GEMM} to achieve high hardware efficiency.

\section{Conclusion}
We propose \method, an accurate and efficient post-training quantization method to enable lossless 8-bit weight and activation quantization for LLMs up to 530B parameters.
\method enables the quantization for both weight and activations for all \texttt{GEMM}s in the LLMs, which significantly reduces the inference latency and memory usage compared with the mixed-precision activation quantization baseline. We integrate \method into PyTorch and FasterTransformer, getting up to 1.56$\times$ inference acceleration and halving the memory footprint. 
\method democratizes the application of LLMs by offering a turnkey solution to reduce the serving cost.

\section*{Acknowledgements}
We thank MIT-IBM Watson AI Lab,  
MIT AI Hardware Program,
Amazon and MIT Science Hub, 
NVIDIA Academic Partnership Award,  
Qualcomm Innovation Fellowship, 
Microsoft Turing Academic Program, and NSF for supporting this research. We thank Haotian Tang, Aohan Zeng, Eric Lin and Jilei Hou for the helpful discussions.

\bibliography{main}
\bibliographystyle{icml2023}

\clearpage
\appendix
\section{Discussion on Weight-Only Quantization}
\label{sec:supp_gptq_compare}
In this work, we study W8A8 quantization so that we can utilize INT8 GEMM kernels to increase the throughput and accelerate inference. There is another line of work that only quantizes the weight of LLMs (\eg, GPTQ~\cite{frantar2022gptq}). It converts the quantized weights to FP16 on the fly for matmul during inference and can also lead to speed up due to the reduced data loading, especially for the generation stage with batch size 1. 

We mainly compare our method with existing work on weight-activation quantization (\ie, W8A8) like~\cite{dettmers2022llmint8, zeroquant, outlier_suppression} since they are under the same setting. Here we would like to give a short discussion about the weight-only quantization methods in LLM settings:
\begin{enumerate}
    \item Firstly, we were trying to compare our method with GPTQ~\cite{frantar2022gptq} but found it difficult due to different implementations. GPTQ's low-bit kenerl~\footnote{\url{https://github.com/IST-DASLab/gptq}} only supports the generation stage with batch size 1 (\ie, only processing a single token at a time), and cannot support the context stage (widely used in different downstream tasks and chatbot) or batch-based setting. Furthermore, its low-bit kernel optimization only targets the OPT-175B model (as stated in the README). At the same time, our work utilizes FasterTransformer for serving large models, which may lead to an unfair advantage if we make a direct comparison.
    \item GPTQ may perform better at handling a small number of input tokens (1 in its experiments) since the process is highly memory-bounded. In contrast, \method may serve better with a batching setting or for the context stage (\ie, when the number of processed tokens is more significant). Nonetheless, some work shows that in production, we can improve the throughput of serving GPT models by 37$\times$ at similar latency with advanced batching~\cite{yu2022orca}. We believe in production, batching will be the future standard, and \method will bring further improvement, even for the generation stage. 
    \item Applications like chatbots need to handle a long context length and potentially run under a batch setting. Due to the two factors, the memory size of the KV cache can no longer be ignored (as shown in~\cite{pope2022efficiently}, the KV cache totals 3TB given batch size 512 and context length 2048, which is 3$\times$ larger than the model weights). In this case, quantization of activation can also help reduce the memory cost from storing the KV cache.  
    \item Finally, we think the two settings are somewhat orthogonal. We believe we can integrate GPTQ's method for a better weight quantization and potentially achieve W4A4 quantization, which will lead to even better hardware efficiency (INT4 instructions are supported on NVIDIA's Hopper GPU architecture). We leave this exploration to future work.
\end{enumerate}

\end{document}